%% file: colm2024_conference.tex
\documentclass{article} 
\usepackage{colm2024_conference}
\input{math_commands}

\usepackage{microtype}
\usepackage{hyperref}
\usepackage{url}
\usepackage{booktabs}
\usepackage{graphicx}
\usepackage{wrapfig}
\usepackage{mathtools}
\usepackage{amsmath}
\usepackage{breqn}
\usepackage{latexsym}
\usepackage{lipsum}
\usepackage{enumerate}
\usepackage{enumitem}
\usepackage{algorithm}
\usepackage{algpseudocode}
\usepackage{bbm}
\usepackage{multirow}

\definecolor{darkblue}{rgb}{0, 0, 0.5}
\hypersetup{colorlinks=true, citecolor=darkblue, linkcolor=darkblue, urlcolor=darkblue}
\definecolor{darkpastelred}{rgb}{0.76, 0.23, 0.13}
\definecolor{coolgrey}{rgb}{0.55, 0.57, 0.67}
\definecolor{lowyellow}{RGB}{241, 196, 15}

\makeatletter
\newcommand*{\rom}[1]{\expandafter\@slowromancap\romannumeral #1@}
\makeatother

\title{Unlocking Temporal Question Answering for Large Language Models with Tailor-Made Reasoning Logic}

\author{
Xingxuan Li$^{1,2}$\thanks{Xingxuan Li is under the Joint Ph.D. Program between DAMO Academy and Nanyang Technological University.}~~
Liying Cheng$^{1}$~~
Qingyu Tan$^{1,3}$~~
\textbf{Hwee Tou Ng}$^3$~~
\textbf{Shafiq Joty}$^{2,4}$\\
\textbf{Lidong Bing}$^{1,5}$\\
$^1$DAMO Academy, Alibaba Group, Singapore ~ $^2$Nanyang Technological University\\
$^3$National University of Singapore ~~ $^4$Salesforce AI\\
$^5$Hupan Lab, 310023, Hangzhou, China\\
\{xingxuan.li, liying.cheng, l.bing\}@alibaba-inc.com\\
\{srjoty\}@ntu.edu.sg ~
\{qtan6, nght\}@comp.nus.edu.sg
}

\colmfinalcopy 
\begin{document}

\maketitle

\input{sections/0_abstract}

\input{sections/1_intro}

\input{sections/2_relatedwork}

\input{sections/3_method}

\input{sections/4_experiments}

\input{sections/5_analysis}

\input{sections/6_conclusion}

\bibliography{colm2024_conference, custom}
\bibliographystyle{colm2024_conference}

\input{sections/9_appendix}

\end{document}

%% file: math_commands.tex
\usepackage{amsmath,amsfonts,bm}

\newcommand{\ie}{{\em i.e., }}
\newcommand{\eg}{{\em e.g., }}

\definecolor{myblue}{RGB}{6, 82, 221}
\definecolor{myorange}{RGB}{211, 84, 0}
\definecolor{lowblue}{RGB}{102,178,255}
\definecolor{justblue}{RGB}{84, 160, 255}
\definecolor{mypurple}{RGB}{108, 92, 231}
\definecolor{mygray}{RGB}{158, 158, 158}
\definecolor{lowpurple}{RGB}{204,153,255}
\definecolor{lowwhite}{RGB}{255,255,255}
\definecolor{verylowpurple}{RGB}{255,102,102}
\definecolor{embcolor}{RGB}{255,255,255}
\definecolor{myred}{RGB}{235, 47, 6} 
\definecolor{mygreen}{RGB}{162, 217, 206} 
\definecolor{fontgrey}{RGB}{44, 62, 80}
\definecolor{lowpurple}{RGB}{210, 180, 222}
\definecolor{mypumpkin}{RGB}{229, 152, 102}
\definecolor{lowgreen}{RGB}{171, 235, 198}
\definecolor{lowgreen2}{RGB}{186, 220, 88}
\definecolor{lowred}{RGB}{245, 183, 177}
\definecolor{lowyellow}{RGB}{241, 196, 15}
\definecolor{mypink}{RGB}{255, 118, 117}
\definecolor{bluemartina}{RGB}{18, 203, 196}
\definecolor{puffin}{RGB}{250, 152, 58}
\definecolor{grass}{RGB}{0, 148, 50}
\definecolor{cnngray}{RGB}{116, 125, 140}
\usepackage{color, colortbl}
\definecolor{Gray}{gray}{0.9}

\def\eqref#1{equation~\ref{#1}}

\def\1{\bm{1}}

\DeclareMathAlphabet{\mathsfit}{\encodingdefault}{\sfdefault}{m}{sl}
\SetMathAlphabet{\mathsfit}{bold}{\encodingdefault}{\sfdefault}{bx}{n}



%% file: sections/0_abstract.tex
\begin{abstract}
The temporal aspect is a significant dimension of our reality.
We notice the challenge that large language models (LLMs) face when engaging in temporal reasoning. Our preliminary experiments show that methods involving the generation of intermediate reasoning steps, such as chain-of-thought and program-aided language models, do not consistently boost the performance of complex temporal question-answering tasks.
This limitation can be attributed to the LLMs' inadequate understanding of temporal information.
To address this problem, we propose TempLogic, a novel framework designed specifically for temporal question-answering tasks across three levels of reasoning. 
TempLogic incorporates retrieval-guided context distillation, temporal data extraction, and tailor-made logic reasoning.
Extensive experiments and analysis demonstrate the effectiveness of our framework in solving intricate time-bound reasoning tasks
\footnote{We will make our code and data publicly available.}.
\end{abstract}

%% file: sections/1_intro.tex
\section{Introduction}
\label{sec:intro}
The ongoing advances in natural language processing (NLP) have paved the way for the emergence of large language models (LLMs), which are now being utilized extensively in various applications.
Models such as GPT-4 \citep{openai2023gpt4}, Vicuna \citep{vicuna2023}, and Alpaca \citep{alpaca} have demonstrated impressive language understanding and generation capabilities and are being used from automated content creation to chatbots \citep{openai2023gpt4, vicuna2023, alpaca}. In utilizing these LLMs, recent work such as chain-of-thought (CoT) \citep{wei2023chainofthought} has been found to further improve performance for tasks that require complex reasoning, such as math problems and symbolic question-answering tasks. 

\begin{wrapfigure}{R}{0.5\textwidth}
\centering
\includegraphics[width=1\linewidth]{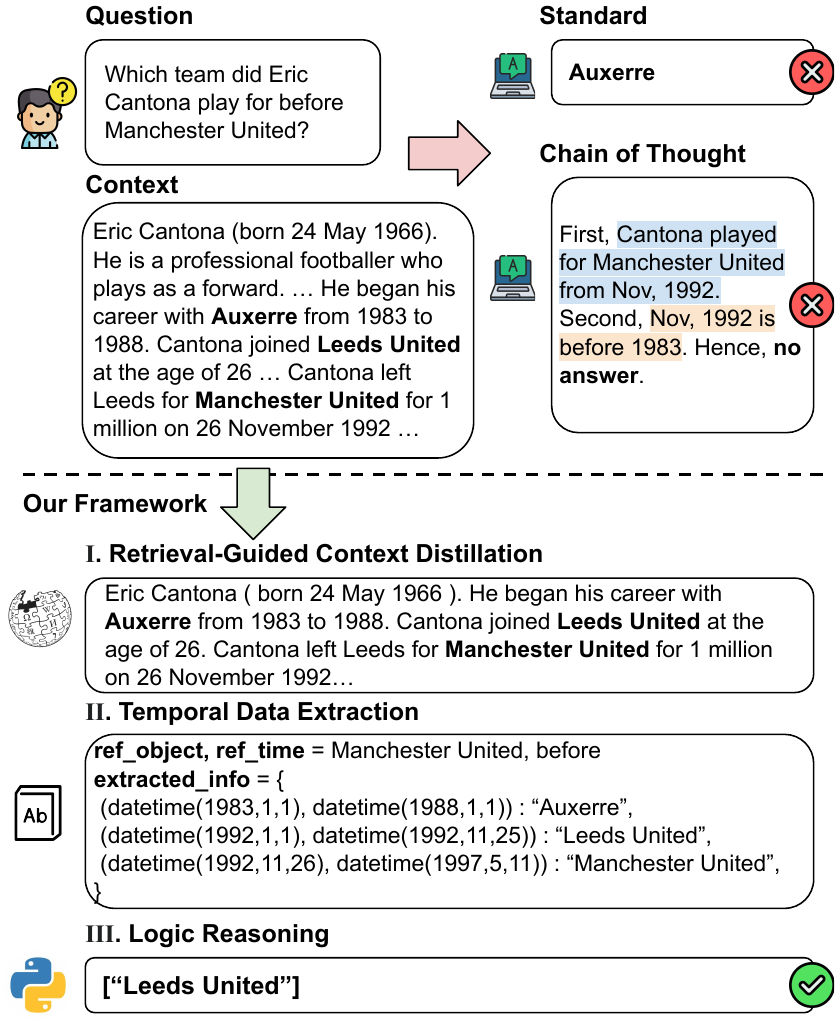}
\caption{Comparison between TempLogic and previous methods. 
TempLogic incorporates three stages: (\rom{1}) retrieval-guided context distillation, (\rom{2}) temporal data extraction, and (\rom{3}) logic reasoning.
}
\label{fig:intro}
\end{wrapfigure}

However, there is a continuing challenge that LLMs face when it comes to temporal reasoning -- the capability to understand and process information that involves time-based concepts and sequences \citep{wei2023chainofthought, zhao2023verifyandedit, chowdhery2022palm}. Though CoT leverages intermediate reasoning steps to guide the generation of the final answer, our investigation reveals that these approaches often fail on the temporal question-answering tasks. 
Figure \ref{fig:intro} provides an example of such a failure. The second reasoning step in the CoT method states that ``Nov, 1992 is before 1983.'', which is incorrect. Consequently, this faulty reasoning in the second step leads to an erroneous answer.

Program-aided language models (PAL) \citep{gao2023pal} proposed a novel approach for addressing temporal-related tasks, wherein it transforms a provided question into Python code and then executes the code to obtain the final answer. 
However, our experimental results indicate that PAL is limited in two ways. First, PAL lacks control over the generation of the code. This means that the generated script may have flawed or even non-executable logic, especially when dealing with complex tasks that require multi-hop reasoning. Second, when utilizing program code to solve temporal problems, it is necessary to extract temporal-related information from the question and context into structured data as the program input. This process becomes challenging when dealing with lengthy contexts, as LLMs face challenges in comprehending the temporal logic within the natural text. Consequently, key information may be omitted from the extracted data leading to inaccurate answers during the execution of the code.


In fact, temporal reasoning tasks can be categorized into three levels: time-time (L1) relation (\eg ``What is the year after 2010?''), time-event (L2) relation (\eg ``What team did Eric Cantona play for in 1995?''), and event-event (L3) relation (\eg ``What team did Eric Cantona play for before Manchester United?'') \citep{tan2023towards}. Each of these relations has its own set of reasoning patterns for solving them. 
In Figure \ref{fig:intro}, we use the above L3 question as an illustration.
To answer this question, we first need to determine the time period during which Cantona played for Manchester United, which spans from 26 November 1992 to 11 May 1997. Next, we compare this time period with all the teams he played for throughout his career. By doing so, we discover that the last time period of his career before 26 November 1992 is from 1 January 1992 to 25 November 1992, during which he played for Leeds United. Hence, the final answer is Leeds United.

Similarly, L1 and L2 relations can be addressed using their own specific reasoning processes. Therefore, we establish the reasoning logic for each relation using Python programming language. This approach ensures that these three sets of reasoning logics encompass all temporal question-answering tasks assuming perfect information is available (\eg having knowledge of all the teams and time periods Cantona played for throughout his career). 
By adopting this approach, we effectively overcome one limitation of PAL, \ie generating scripts that may contain errors and logic flaws.
Note that establishing the precise reasoning behind each relation is not a time-consuming task, and the benefits it yields are obviously worthwhile.

Another difficulty is that extracting accurate and complete information from the context into structured input data for the reasoning logic poses a challenge for LLMs, especially when dealing with lengthy contexts.
For example in Figure \ref{fig:intro}, where in stage \rom{2}, the objective is to extract a dictionary containing all the teams Cantona played for throughout his career. The entry ``(datetime(1992,1,1), datetime(1992,11,25): ``Leeds United'')'' can be inferred from the following three sentences: ``Eric Cantona (born 24 May 1966).'', ``Cantona joined Leeds United at the age of 26.'' and ``Cantona left Leeds for Manchester United for 1 million on 26 November 1992.''. When there are numerous redundant sentences between these three key sentences, LLMs often fail to extract such information. 
To tackle this problem, we perform retrieval-guided context distillation. This process involves utilizing a retrieval system to retrieve relevant knowledge related to the question. Subsequently, we distill the original context by incorporating the retrieved knowledge. By employing this approach, we remove redundant sentences (\eg ``He is a professional footballer who plays as a forward.'') and retain only the sentences relevant to the question. With the distilled context, it is easier for LLMs to comprehend and extract temporal data into a dictionary.

\begin{figure*}[!t]
\centering
\includegraphics[width=0.99\linewidth]{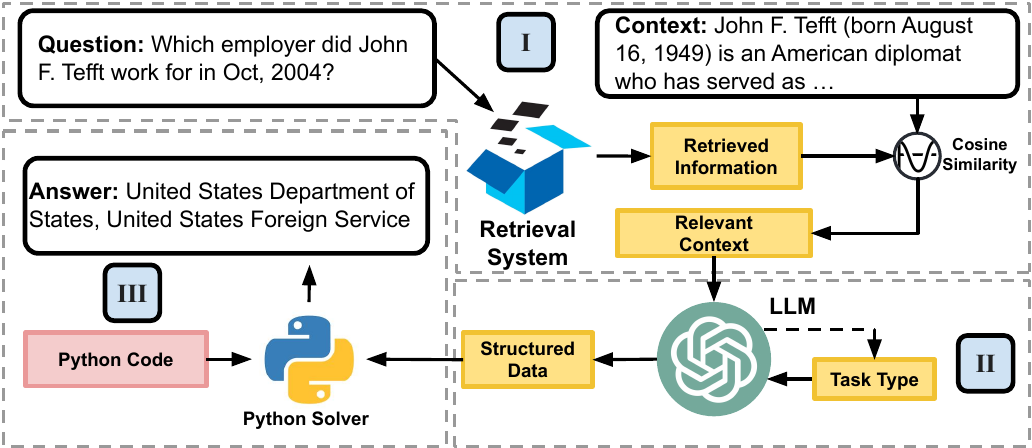}
\caption{An overview of our proposed framework.
\textbf{(\rom{1}) Retrieval-guided context distillation.} The question was inputted into a retrieval system to obtain relevant information. The retrieved information is then used to distill the most relevant sentences from the raw context.
\textbf{(\rom{2}) Temporal data extraction.} Using the distilled context, an LLM is utilized to extract a temporal dictionary.
\textbf{(\rom{3}) Logic Reasoning.} The final answer is obtained by combing and executing the extracted dictionary and a tailor-made reasoning logic. 
}
\label{fig:framework}
\end{figure*}

On the whole, we propose TempLogic, a three-stage framework designed to support LLMs in solving complex temporal question-answering tasks.
As shown in Figure \ref{fig:framework}, our framework incorporates tailor-made reasoning logic that addresses all three relations of temporal reasoning. Furthermore, we perform retrieval-guided context distillation to aid LLMs in effectively extracting temporal data from the context into a Python dictionary. TempLogic combines the extraction capability of LLMs and the logical reasoning capability of a runtime (\eg Python interpreter).

In summary, our main contributions are:
(1) Our preliminary experiments reveal that generating intermediate reasoning steps does not consistently boost the performance of complex temporal question-answering tasks.
(2) We propose a novel framework TempLogic incorporating tailor-made reasoning logics that address all three relations of temporal reasoning. TempLogic combines the extraction capability of LLMs and the logical reasoning capability of a runtime while performing retrieval-guided context distillation. 
(3) We first use a stricter exact match (SEM) score for evaluating temporal question-answering tasks. SEM offers a more suitable evaluation for multiple-answer questions, a scenario often encountered but not appropriately addressed by existing metrics.
(4) Extensive experiments and ablation studies on two benchmark datasets verify the effectiveness of our framework.

%% file: sections/2_relatedwork.tex
\section{Related Work}
\subsection{Temporal Question Answering}
In recent years, multiple temporal question-answering datasets have been proposed. 
The first line of temporal reasoning datasets works on temporal QA over knowledge graphs (KGs), such as TempQuestions \cite{Zhen2018TempQuestionsAB}, TimeQuestions \cite{Jia2021ComplexTQ}, TEQUILA \cite{Jia2018TEQUILATQ}, and C\textsc{ron}Q\textsc{uestions} \cite{Saxena2021QuestionAO}. The task of KGQA asks the model to rank all entities in a knowledge graph for each temporal query. However, this line of works presumes that all entities are known to the system and cannot perform temporal reasoning solely based on natural text.

In this work, we focus on studying the temporal reasoning of large language models. There have been several datasets proposed for temporal question answering, such as SituatedQA \cite{zhang-choi-2021-situatedqa} and StreamingQA \cite{Livska2022StreamingQAAB}. These two datasets aim to answer open-domain time-sensitive questions under both open-book and closed-book settings.
T\textsc{emp}LAMA \cite{dhingra-etal-2022-time} is proposed to answer closed-book temporal questions. ArchivalQA \cite{wang2022archivalqa} is designed for temporal news open-domain QA. Unlike the previous datasets that focus on either open-book QA using retrieval-based models or closed-book approaches that only rely on the knowledge stored in the model's parameters, TimeQA \cite{chen2021dataset} and TempReason \cite{tan2023towards} are two temporal QA datasets that focus on the reasoning aspect of the temporal QA task.
Therefore, we conduct our experiments on these two datasets in this paper.

\subsection{Large Language Models}
In-context learning with large language models (LLMs) like InstructGPT \citep{ouyang2022training} and LLaMA \citep{Touvron2023LLaMAOA} has proven successful in numerous tasks \citep{zhao2023verifyandedit, ding2022gpt3, li2023does}. 
Methods such as chain-of-thought (CoT) \citep{wei2023chainofthought} further improve the performance of LLMs on reasoning tasks by generating intermediate reasoning steps. 
However, the CoT method continues to struggle with accuracy in many complex reasoning tasks, including arithmetic computation and symbolic reasoning \citep{gao2023pal}. \citet{gao2023pal} proposes program-aided language models (PAL), which aim to solve these issues by offloading calculations and part of the reasoning process to a Python interpreter. 
Their approach shares similarities with ours on a conceptual level. However, unlike our method, PAL does not perform well in intricate temporal question-answering tasks. In contrast, we demonstrate the efficacy of our framework in time-bound tasks of varying difficulty levels. Additionally, while PAL uses LLMs to generate Python code, potentially leading to hallucinations, we only utilize the information extraction capabilities of LLMs in tandem with the problem-solving capabilities of a Python solver to address the problem more accurately. Essentially, we combine the best of both worlds.

%% file: sections/3_method.tex
\section{Framework}
To tackle the intricate temporal question-answering tasks, we design a straightforward yet effective end-to-end framework called TempLogic. As demonstrated in Figure \ref{fig:framework}, our framework consists of three stages: (1) retrieval-guided context distillation, (2) temporal data extraction, and (3) logic reasoning. 


\subsection{Retrieval-Guided Context Distillation}
\label{sec:step1}

Lengthy context poses challenges for large language models (LLMs), hindering their ability to extract structured knowledge effectively. Therefore, our initial step involves retrieval-guided context distillation. This step incorporates two essential components: knowledge retrieval and context distillation.

\paragraph{Knowledge Retrieval}

We define a set of test examples as $\{t_1,t_2,...,t_n\}\in T$.
Additionally, we define the question and its corresponding context for $t_i$ as $q_i$ and $c_i$, respectively. By utilizing an external retrieval system $R(\cdot)$, we acquire a set of retrieved documents $\{d^1_i,d^2_i,...,d^m_i\}\in D_i$ for $q_i$ as
\begin{equation}
    D_i = R(q_i) \text{.}
\end{equation}

\input{tbls/code}

\paragraph{Context Distillation}
Given that the accurate answer $a_i$ is contained within the original context $c_i$, our task is to extract the most pertinent sentences from $c_i$ using the guidance of $D_i$ during the distillation process.
We first employ a sentence embedding model $e(\cdot)$ to generate the sentence embedding for each sentence $c^k_i \in c_i$ resulting in $e(c^k_i) \in E_i$.
Additionally, for each retrieved sentence $d^j_i \in D_i$, we compute its corresponding sentence embedding $e(d^j_i)$. To select the most relevant sentences for $d^j_i$, we calculate the cosine similarity between $e(d^j_i)$ and $E_i$. From this similarity calculation, we choose the top three $e(c^k_i)$ with the highest cosine similarity to $e(d^j_i)$. These selected $c^k_i$ sentences serve as the distilled sentences for $d^j_i$. By repeating these steps for all $d^j_i \in D_i$, we compile all unique distilled sentences and concatenate them to form the final distilled context $s_i$.
This approach allows us to eliminate redundant sentences, such as the sentence ``He is a professional footballer who plays as a forward.'' in Figure \ref{fig:intro}. As a result, this process makes it easier and more efficient to extract structured information in subsequent stages.



\subsection{Temporal Data Extraction}
Since the three levels of temporal questions demand varied temporal details, the temporal data extraction stage involves two critical steps: task classification and data extraction.
\paragraph{Task classification}
We employ in-context learning (ICL) to classify the questions into three distinct tasks: L1, L2, and L3.
We provide one demonstration example for each task to create a 3-shot ICL setting.
Due to the simplicity of the classification, the accuracy of the classification is close to 100\%.

\input{tbls/step1}
\paragraph{Data extraction}
In this step, we utilize the extraction capability of LLMs to obtain a structured temporal dictionary (extracted\textunderscore info) from the context and other necessary information from the question to facilitate logical reasoning in the later stage. 
More specifically, the necessary information from the question includes the object (ref\textunderscore obj) and the reference time relationship (ref\textunderscore time). 
An example can be found in Figure \ref{fig:intro}, where the object is ``Manchester United'' and the reference time relationship is ``before''. We design a one-shot example $P_{train}$, which consists of training input (\ie question and context), and training output (\ie extracted\textunderscore info, ref\textunderscore obj, and ref\textunderscore time) as shown in Table \ref{tab:step1_prompt}\footnote{Details of the one-shot example can be found in \ref{sec:sir}.}. The prompt $P_i$ for test example $t_i$ is therefore formed by the training prompt $P_{train}$, test question $q_i$, and test context $c_i$.

We obtain $D_i$, $r_{o,i}$, and $r_{t,i}$ as
\begin{equation}
    D_i, r_{o,i}, r_{t,i} \sim M_{\tau}(P_i) \text{,}
\end{equation}
where $D_i$, $r_{o,i}$, and $r_{t,i}$ are the extracted\textunderscore info, ref\textunderscore obj, and ref\textunderscore time for $t_i$, respectively.
$M_{\tau}(\cdot)$ is the LLM with $\tau$ as the temperature used during the decoding process\footnote{We use $\tau=0$ for all experiments for the sake of reproducibility.}.

\subsection{Logic Reasoning}

Upon extracting $D_i$, $r_{o,i}$, and $r_{t,i}$ from the previous stage, we proceed by integrating them with our tailor-made reasoning logic. To illustrate the design, we use the L3 relation as an example and present the corresponding code design in Algorithm \ref{alg:l3code}. 
We first define a function called \uppercase{{time\textunderscore relation}}, which takes the start and end times of two time periods as input and determines their relation. The possible relations include ``after'', ``before'', ``contains'', ``contained by'', ``simultaneous'', or ``overlaps''.
Next, we create a dictionary $D'_i$ where the object serves as the key, and the corresponding time period is the value.
As mentioned earlier in Section \ref{sec:intro}, the first step to address an L3 reasoning task is to determine the time period of the $r_{o,i}$. By following the algorithm, we acquire the start time $r_{st,i}$ and end time $r_{et,i}$ of the reference object:
\begin{equation}
    r_{st,i}, r_{et,i} = D'_i(r_{o,i}).
\end{equation}
For each object $k'_{i,j}$ in $D'_i$ that is not $r_{o,i}$ itself, we utilize the \uppercase{time\textunderscore relation} function to compare the two relations. If they match, we consider $k'_{i,j}$ to be a correct answer.
The reasoning logic for L1 and L2 relations can be found in Appendix \ref{app:logic}.
By implementing the reasoning logic using the Python programming language, we execute the script and obtain the final answer.



%% file: tbls/code.tex
\begin{algorithm*}
    \caption{L3 Reasoning Logic}
    \label{alg:l3code}
    \begin{algorithmic}
        \Require $\textrm{Reference Object } r_o$, $\textrm{Reference Time Relation } r_t$, $\textrm{Temporal Data Dictionary } D$;
        \Ensure $\textrm{Reversed Temporal Data Dictionary } D'$, $\textrm{List of answers } a$;
        \Procedure{time\textunderscore relation}{$st_1$,$et_1$,$st_2$,$et_2$} \Comment{Function to determine the relation of two time periods}
            \If{$et_1 \leq st_2$}
                \Return{after}
            \ElsIf{$et_2 \leq st_1$}
                \Return{before}
            \ElsIf{$st_2 < st_1$ and $et_2 > et_1$}
                \Return{contains}
            \ElsIf{$st_1 < st_2)$ and $et_1 > et_2$}
                \Return{contained by}
            \ElsIf{$st_1 = st_2$ and $et_1 = et_2$}
                \Return{simultaneous}
            \Else{}
                \Return{overlaps}
            \EndIf
        \EndProcedure
        \For{$k$,$v$ in $D$} \Comment{Reverse the key to object and value to time period}
            \State $D'(v) \gets k$
        \EndFor

        \State $r_{st},r_{et} \gets D'(r_o)$ \Comment{Start time and end time of the reference object}
        \For{$k'$,$v'$ in $D'$, $k' \neq r_o$} \Comment{Check the time period relation of the reference object and other objects}
            \If{$r_t =$ \Call{time\textunderscore relation}{$r_{st}$,$r_{et}$,$v'_{st}$,$v'_{et}$}} \Comment{Add the object to the answer if time period matches}
                \State \textbf{insert} $k'$ \textbf{into} $a$
            \EndIf
        \EndFor        
    \end{algorithmic}
\end{algorithm*}

%% file: tbls/step1.tex
\begin{wraptable}{R}{0.45\textwidth}
    \centering
    \resizebox{\linewidth}{!}{
	\setlength{\tabcolsep}{0mm}{
            \begin{tabular}{p{8cm}}
            \toprule
                Instruction: Extract information from the question and context. Strictly follow the below example. \\
                Question: \textcolor{myblue}{[Train Question]}\\
                Context: \textcolor{myblue}{[Train Context]}\\
                extracted\textunderscore info = \textcolor{myblue}{[Train extracted\textunderscore info]}\\
                ref\textunderscore obj = \textcolor{myblue}{[Train ref\textunderscore obj]}\\
                ref\textunderscore time = \textcolor{myblue}{[Train ref\textunderscore time]}\\\\
                
                Question: \textcolor{myred}{[Test Question]}\\
                Context: \textcolor{myred}{[Test Context]}\\
                extracted\textunderscore info = \\
            \bottomrule
            \end{tabular}
        }
    }
    \caption{
        Prompt for temporal data extraction stage in our framework. 
        Text in \textcolor{myblue}{blue}: the specific question, context, extracted\textunderscore info, ref\textunderscore obj, and ref\textunderscore time of one-shot example.  
        Text in \textcolor{myred}{red}: the specific question and context for test data.
    }
    \label{tab:step1_prompt}
\end{wraptable}

%% file: sections/4_experiments.tex
\section{Main Experiments}
\subsection{Datasets}
We evaluate our method on temporal question-answering tasks.

\paragraph{Comprehensive Temporal Reasoning Benchmark (TempReason)} TempReason \citep{tan2023towards} is a dataset that consists of temporal question-answering tasks classified into three levels, namely L1, L2, and L3. 
L1 questions follow a ``time-time'' structure, such as ``What is the year after 2010?''.
L2 questions are designed around an ``event-time'' structure, such as ``What team did Eric Cantona play for in 1995?'' 
Lastly, L3 questions follow an ``event-event'' structure, such as ``What team did Eric Cantona play for before Manchester United?'' 
It is important to note that the L1 questions are the easiest, and the L3 questions are generally more challenging compared to the L2 questions, due to their inherent complexity and relational nature.
For both L2 and L3 levels, factual context is supplied to aid in deriving the answers.
Factual context takes the form of well-structured temporal data, for instance, ``Eric Cantona played for Manchester United from Nov, 1992 to May, 1997.'' This kind of well-structured context is more readily and effectively processed by LLMs for information extraction, thereby aiding the solution generation process. Crucially, the factual context always guarantees the inclusion of the correct answer.

\paragraph{Time-Sensitive Questions (TimeQA)} TimeQA \citep{chen2021dataset} is a dataset designed to evaluate the temporal reasoning ability of models. It presents challenges in two key dimensions: understanding time-based facts and reasoning over these temporal elements. The dataset includes tasks of two complexity levels, namely ``easy'' and ``hard''. For the ``easy'' tasks, the facts are clear-cut and without any instances of overlapping time periods. This means that the information necessary for answering these questions is explicit. On the other hand, the ``hard'' tasks require deeper analysis, as the context of these questions often contains implicit facts. 
The context provided in TimeQA poses a higher degree of difficulty due to its direct extraction from Wikipedia, a source that lacks the well-structured formatting characteristic of TempReason.
Additionally, we post-process the context of TimeQA in a similar format as TempReason's factual context. However, the derived factual contexts contain both yearly and monthly data, which may not be as accurate as TempReason since all factual contexts in TempReason are monthly data.
In our experiments, we evaluate both the raw context from the original dataset and the post-processed factual context. Crucially, both contexts always guarantee the inclusion of the correct answer. The questions in TimeQA are L2 level reasoning.

\subsection{Baselines}
To provide a more comprehensive overview of where our framework stands, we compare with both standard and chain-of-thought (CoT) prompting \citep{ouyang2022training, wei2023chainofthought}. For CoT, we undertake a comprehensive analysis with various prompting orders. Furthermore, we compare with the state-of-the-art method program-aided language models (PAL) \citep{gao2023pal}. Details can be found in Appendix \ref{app:baselines}.

\subsection{Experimental Setup}
Both the TempReason and TimeQA datasets consist of questions that may have either single or multiple answers. To assess the performance in each case, we examine these two scenarios independently. We randomly select 500 data points from the TempReason L2 category for both single- and multiple-answer questions. As the L1 and L3 category does not include questions with multiple answers, we only sample 500 data points for single-answer questions. For TimeQA, we select 100 data points each for single- and multiple-answer questions due to limited multiple-answer questions in the dataset. Since the focus of this paper is on temporal reasoning, we adopt the ReasonQA problem setting proposed in \citet{tan2023towards} in our experiments.

\input{tbls/main_exp}

For all our experiments, we employ the widely used version of InstructGPT (\texttt{text-davinci-003}) as the model.
We set the temperature to 0 to ensure the reproducibility of our experiments. 
We decided not to use chat models such as ChatGPT and GPT-4, as they demonstrate inconsistent results across multiple runs, even when we use the frozen snapshot with the temperature set at 0. We hypothesize that, while the model parameters remain static, the tokenizer may be subject to changes over time, leading to differing outcomes. Given the sensitivity of temporal information to tokenization and our commitment to reproducibility, we opt not to include ChatGPT in our experiments. For the retrieval model, we utilize DrQA \citep{drqa}. For the sentence embedding model, we use SimCSE \citep{simcse}.
Prior efforts of temporal question answering (\citealp{chen2021dataset}; \citealp{dhingra-etal-2022-time}; \citealp{Livska2022StreamingQAAB}) followed the evaluation protocol of the SQuAD 2.0 benchmark \cite{rajpurkar-etal-2018-know}, using exact match and token-level F1 score. However, these two metrics (EM and token-F1) are not suitable for evaluating questions with multiple answers, because the SQuAD benchmark takes the max score for all the possible answers. However, for the temporal QA task, there are many cases where multiple answers are valid for a given temporal query. For example, executive officials may have multiple positions in different companies. To this end, we first define the strict exact match (\textbf{SEM}) score. Predictions will only be considered correct if all the gold answers are matched for a given question. We also evaluate our methods by answer-level F1 score (\textbf{F1}), which is a stricter metric compared to token-level F1 score.

\subsection{Results on TempReason}
\label{sec:tempreason}

We present the experimental results for TempReason in Table \ref{tb:tempreason}. CoT (Q+C+R+A) refers to a CoT method with the order of question, context, reasoning steps, and final answers in the prompt. We have the following observations: 
\textbf{(a)} Our method significantly outperforms standard and CoT baselines on all three tasks, as well as on single- and multiple-answer question tasks. 
Specifically, our method enhances the performance on L2 Single by a remarkable 21.2\% over the standard prompting method. Even more notable improvements are observed for L3 Single and L2 Multi, where the performance is boosted by 39\% and 32.17\% respectively. The performance boosts across all three levels of tasks benefit from our tailored-made reasoning logics and extraction of temporal data.
\textbf{(b)} The effectiveness of CoT methods can significantly vary based on the order of elements in the prompt, which matches the observation made by \citet{ye2022unreliability}. For example, the SEM of CoT (Q+C+A+R) on L2 Multi is only 0.2\%, while that of CoT (C+Q+R+A) is 19.8\%. 
\textbf{(c)} The CoT methods do not outperform the standard prompting method. We attribute this to the errors in the intermediate reasoning steps, which lead to incorrect answers ultimately. Further analysis on this issue will be presented in Section \ref{sec:cotworse}.
\textbf{(d)} On average, our method achieves a less significant performance improvement of 2.3\% over PAL. This is mainly due to the well-formatted factual context in TempReason. With such context, PAL is capable of extracting temporal dictionaries without any explicit context processing.

\input{tbls/timeqa}
\subsection{Results on TimeQA}
We present the experimental results for TimeQA in Table \ref{tb:timeqa}.
We evaluate both original context and factual context. And we have the following observations:
\textbf{(a)} TempLogic significantly outperforms all Standard and CoT baselines across all tasks by 13.38\% and 19.59\% on average.
\textbf{(b)} We observe that CoT (Q+C+A+R) performs poorly with multiple-answer questions. CoT (C+Q+A+R) also underperforms in ``hard'' multiple-answer questions. This shows that prioritizing reasoning before answering significantly enhances performance \citep{wei2023chainofthought}.
\textbf{(c)} Similar to TempQA, our framework achieves a less significant average improvement of 4.25\% compared to PAL when utilizing the factual context. However, when utilizing the raw context, our framework achieves a significant average improvement of 14.5\% over PAL. This underscores the substantial role played by knowledge-guided context distillation in driving the improvement. By combining retrieval-guided context extraction and tailor-made logical reasoning, TempLogic guarantees accuracy in solving temporal question-answering tasks.

%% file: tbls/main_exp.tex
\begin{wraptable}{R}{0.55\textwidth}
    \centering
    \resizebox{\linewidth}{!}{
        \begin{tabular}{cccccc}
\toprule
\multirow{2}{*}{\textbf{Method}} & \textbf{L2 Single} & \textbf{L2 Single} & \textbf{L3 Single} & \multicolumn{2}{c}{\textbf{L2 Multi}} \\ 
\cmidrule(lr){2-2}
\cmidrule(lr){3-3}
\cmidrule(lr){4-4}
\cmidrule(lr){5-6}
                                     & \textbf{SEM} & \textbf{SEM}        & \textbf{SEM}        & \textbf{SEM}       & \textbf{F1}       \\ 
                                     \midrule
\multicolumn{1}{c}{Standard}      &72.20     & 72.60              & 52.60              & 21.00             & 59.38             \\
\multicolumn{1}{c}{CoT (Q+C+R+A)} &-  & 71.40              & 51.80              & 16.20             & 43.19             \\
\multicolumn{1}{c}{CoT (C+Q+R+A)} &-  & 68.40              & 38.80              & 19.80             & 47.08             \\
\multicolumn{1}{c}{CoT (Q+C+A+R)} &-  & 64.00              & 33.00              & 0.20              & 42.99             \\
\multicolumn{1}{c}{CoT (C+Q+A+R)} &-  & 60.60              & 54.60              & 2.20              & 48.21             \\ 
\multicolumn{1}{c}{CoT (avg.)} &78.60  & 66.10              & 44.55              & 9.60              & 45.37             \\ 
\multicolumn{1}{c}{PAL} & \underline{96.40}  & \underline{91.20} & \underline{89.40} & \underline{73.20}  & \underline{86.93}             \\ 
\midrule
\multicolumn{1}{c}{TempLogic} & \textbf{97.60} & \textbf{93.80}              & \textbf{91.60}              & \textbf{76.40}             & \textbf{91.55}             \\ 
\bottomrule
\end{tabular}
    }
    \caption{Experimental results on TempReason.}
    \label{tb:tempreason}
\end{wraptable}

%% file: tbls/timeqa.tex
\begin{table*}[t!]
    \centering
    \resizebox{1\linewidth}{!}{
        \begin{tabular}{ccccccccccccc}
\toprule
\multirow{3}{*}{\textbf{Method}} & \multicolumn{2}{c}{\textbf{Easy Single}} & \multicolumn{4}{c}{\textbf{Easy Multi}} & \multicolumn{2}{c}{\textbf{Hard Single}} & \multicolumn{4}{c}{\textbf{Hard Multi}} \\ 
\cmidrule(lr){2-3}
\cmidrule(lr){4-7}
\cmidrule(lr){8-9}
\cmidrule(lr){10-13}
& \multicolumn{1}{c}{\textbf{F.C.}} & \multicolumn{1}{c}{\textbf{R.C.}} & \multicolumn{2}{c}{\textbf{F.C.}} & \multicolumn{2}{c}{\textbf{R.C.}} & \multicolumn{1}{c}{\textbf{\textbf{F.C.}}} & \multicolumn{1}{c}{\textbf{R.C.}} & \multicolumn{2}{c}{\textbf{F.C.}} & \multicolumn{2}{c}{\textbf{R.C.}} \\
\cmidrule(lr){2-2}
\cmidrule(lr){3-3}
\cmidrule(lr){4-5}
\cmidrule(lr){6-7}
\cmidrule(lr){8-8}
\cmidrule(lr){9-9}
\cmidrule(lr){10-11}
\cmidrule(lr){12-13}
 & \textbf{SEM} & \textbf{SEM} & \textbf{SEM} & \textbf{F1} & \textbf{SEM} & \textbf{F1} & \textbf{SEM} & \textbf{SEM} & \textbf{SEM} & \textbf{F1} & \textbf{SEM} & \textbf{F1} \\ \hline
Standard & 80.00 & 38.00 & 49.00 & 81.24 & 1.00 & \underline{36.89} & 58.00 & \underline{33.00} & 33.00 & 72.59 & 1.00 & 28.66 \\
CoT (Q+C+R+A) & 79.00 & 29.00 & 58.00 & \underline{83.99} & \underline{7.00} & 36.47 & 43.00 & 23.00 & 25.00 & \underline{76.43} & 3.00 & 25.89 \\
CoT (C+Q+R+A) & 79.00 & \underline{43.00} & 29.00 & 73.03 & 6.00 & 31.77 & 54.00 & 28.00 & 44.00 & 67.88 & 0.00 & \underline{31.76} \\
CoT (Q+C+A+R) & 77.00 & 27.00 & 3.00 & 62.05 & 0.00 & 26.02 & 44.00 & 20.00 & 1.00 & 49.37 & 1.00 & 23.6 \\
CoT (C+Q+A+R) & 80.00 & 37.00 & 35.00 & 75.99 & 0.00 & 33.95 & 56.00 & 32.00 & 9.00 & 60.12 & 1.00 & 27.16 \\
CoT (avg.) & 78.75 & 34.00 & 31.25 & 73.77 & 3.25 & 32.05 & 49.25 & 25.75 & 19.75 & 63.45 & 1.25 & 27.10 \\
PAL & \underline{83.00} & 24.00 & \underline{71.00} & 82.32 & 6.00 & 33.85 & \underline{59.00} & 25.00 & \underline{52.00} & 72.93 & \underline{5.00} & 23.16 \\
\midrule
TempLogic & \textbf{84.00} & \textbf{49.00} & \textbf{75.00} & \textbf{89.91} & \textbf{13.00} & \textbf{42.51} & \textbf{67.00} & \textbf{41.00} & \textbf{56.00} & \textbf{76.08} & \textbf{15.00} & \textbf{33.16} \\
\bottomrule
\end{tabular}
    }
    \caption{Experimental results on TimeQA. F.C. stands for factual context. R.C. stands for raw context. }
    \label{tb:timeqa}
\end{table*}

%% file: sections/5_analysis.tex
\section{Analysis and Ablation Studies}

\subsection{CoT Methods Do Not Always Outperform Standard Prompting}
\label{sec:cotworse}
\input{tbls/case_studies}
As mentioned in one of the observations of Section \ref{sec:tempreason}, the CoT methods do not always outperform the standard prompting method in TempReason. In this section, we illustrate a few cases generated by both InstructGPT (\texttt{text-davinci-003}) and GPT-4 (\texttt{gpt-4-0314}). 
As shown in Table \ref{tb:case1}, the standard prompting methods (both InstructGPT and GPT-4) derive the correct answer to the question. However, the CoT methods, even though powered by these powerful LLMs, derive incorrect answers.
The cause of this discrepancy lies in the inaccurate generation of the intermediate reasoning steps by the CoT methods, which ultimately results in an incorrect answer.

Similarly, as shown in an example drawn from the multiple-answer questions from TempReason L2 (\ref{tb:case2}), the CoT methods (both InstructGPT and GPT-4) are unable to answer the question correctly. We also observe that both methods struggle to identify multiple answers. They stop the reasoning steps as soon as the first matching answer is found, indicating a limitation in handling questions that require multiple answers. However, our method is fully capable of handling such instances, given that the Python solver is supplied with accurate and comprehensive extracted information from the question and context.

\subsection{Necessity of External Python Solver}
\input{tbls/ablation_necessity}
The external Python solver serves as a vital component of our framework. 
We conduct ablation studies to examine the implications of utilizing an external Python solver for code execution instead of leveraging the reasoning capability of LLMs via in-context learning.
To ensure a fair comparison, the inputs of both methods are the same, which contain the predefined Python code and all extracted information from previous stages.
Table \ref{tb:ablation_necessity} shows the comparison among the baseline performance of using the standard prompting method, our method with and without executing the code with an external Python interpreter.
It shows a significant drop in performance, from 93.8 to 61.2, for single-target instances, and from 76.4 to 33.8 for multi-target instances when we opt not to execute the code using an external Python solver.
Without executing the code, our method performs worse than the standard prompting baseline on L2 Single, which again demonstrates the necessity of the external Python solver.
This finding aligns with \citet{gao2023pal} as well.

%% file: tbls/case_studies.tex
\begin{table*}[t!]
    \centering
    \scalebox{0.67}{
\begin{tabular}{ll}
\toprule
\textbf{Question}               & Which employer did Richard Tol work for before Economic and Social Research Institute?                                                                                                                                                                                                                                                                                                                                                                                               \\ \hline
\textbf{Context}                & \begin{tabular}[c]{@{}l@{}}Richard Tol works for Vrije Universiteit Amsterdam from Jan, 1992 to Jan, 1992. \\ Richard Tol works for Economic and Social Research Institute from Jan, 2006 to Jan, 2012. \\ Richard Tol works for University of Hamburg from Jan, 2000 to Jan, 2006. \\ Richard Tol works for University of Sussex from Jan, 2012 to Dec, 2022.\end{tabular}                                                                                                          \\ \hline
\textbf{Gold Answer}            & \cellcolor[HTML]{FFFFFF}\textbf{University of Hamburg}                                                                                                                                                                                                                                                                                                                                                                                                                               \\ \hline
\textbf{Standard (InstructGPT)} & \cellcolor[HTML]{FFFFFF}\textcolor{myblue}{\textbf{University of Hamburg}}                                                                                                                                                                                                                                                                                                                                                                                                                               \\ \hline
\textbf{CoT (InstructGPT)}      & \begin{tabular}[c]{@{}l@{}}Answer: \textcolor{myred}{\textbf{Vrije Universiteit Amsterdam}}. Reasoning: \\ First, Richard Tol works for Vrije Universiteit Amsterdam from Jan, 1992 to Jan, 1992. \\ Second, Richard Tol works for University of Hamburg from Jan, 2000 to Jan, 2006. \\ Third, Richard Tol works for Economic and Social Research Institute from Jan, 2006 to Jan, 2012. \\ Therefore, the employer before Economic and Social Research Institute is Vrije Universiteit Amsterdam.\end{tabular} \\ \hline
\textbf{Standard (GPT-4)}       & \cellcolor[HTML]{FFFFFF}\textcolor{myblue}{\textbf{University of Hamburg}}                                                                                                                                                                                                                                                                                                                                                                                                                                        \\ \hline
\textbf{CoT (GPT-4)}            & \begin{tabular}[c]{@{}l@{}}Answer: \textcolor{myred}{\textbf{Vrije Universiteit Amsterdam}}. Reasoning:\\ Richard Tol worked for Vrije Universiteit Amsterdam from Jan, 1992 to Jan, 1992, \\ and then he worked for Economic and Social Research Institute from Jan, 2006 to Jan, 2012.\end{tabular}                                                                                                                                                                                                            \\ \hline
\textbf{TempLogic}                   & \cellcolor[HTML]{FFFFFF}\textcolor{myblue}{\textbf{University of Hamburg}}                                                                                                                                                                                                                                                                                                                                                                                                                                        \\ \bottomrule
\end{tabular}
    }
    \caption{A case study of a L3 single-answer question from TempReason. Answers highlighted in blue are correct, whereas thoses marked in red are incorrect.}
    \label{tb:case1}
\end{table*}

%% file: tbls/ablation_necessity.tex

\begin{wraptable}{R}{0.4\textwidth}
    \centering
\resizebox{1\linewidth}{!}{
\begin{tabular}{ccc}
\toprule
\multirow{2}{*}{\textbf{Method}}       & \textbf{L2 Single} & \textbf{L2 Multi} \\ 
\cmidrule(lr){2-2}
\cmidrule(lr){3-3}
                                           & SEM                & SEM               \\ 
\midrule
\multicolumn{1}{c}{Standard}              & 72.60              & 21.00             \\
\multicolumn{1}{c}{TempLogic (w/o)} & 61.20              & 33.80             \\ 
\multicolumn{1}{c}{TempLogic (w)}   & 93.80              & 76.40             \\
\bottomrule
\end{tabular}
    }
    \caption{Analysis of our method with or without executing the code with an external Python interpreter.}
    \label{tb:ablation_necessity}
\end{wraptable}

%% file: sections/6_conclusion.tex
\section{Conclusions}
\vspace{-0.5em}
Large language models have shown remarkable progress in natural language processing and are extensively used in various applications.
However, complex reasoning tasks such as temporal reasoning pose a challenge for LLMs.
Recent works on intermediate reasoning steps have improved their performance, but it may not always work for temporal reasoning tasks.
To address this issue, in this work, we propose TempLogic, a novel framework that incorporates retrieval-guided data extraction and tailor-made logic reasoning.
Extensive experiments show that TempLogic effectively handle all levels of temporal reasoning tasks.

%% file: sections/9_appendix.tex
\clearpage
\appendix

\section{Reasoning Logics for L1 and L2}
\input{tbls/l1_algo}
\input{tbls/l2_algo}
\label{app:logic}
Reasoning logic for L1 and L2 are shown in Algorithm \ref{alg:l1code} and Algorithm \ref{alg:l2code} respectively.

\section{Experiment Baselines}
\label{app:baselines}
To provide a more comprehensive overview of where our framework stands, we compare with the following baselines:
\paragraph{Standard Prompting (Standard)} Given one-shot training example in the prompt, standard prompting \citep{ouyang2022training} directly predicts the answer.

\paragraph{Chain of Thought Prompting (CoT)} CoT \citep{wei2023chainofthought} generates several intermediate reasoning steps prior to the final answer, aiming to enhance the reasoning capabilities of LLMs on complex reasoning tasks. It has been observed by \citep{ye2022unreliability} that the effectiveness of CoT can be influenced by the sequence of questions (Q), context (C), reasoning steps (R), and final answers (A). As a result, we undertake a comprehensive analysis with various CoT methodologies, taking into account different prompting orders.

\paragraph{Program-aided Language Models (PAL)} PAL \citep{gao2023pal} uses the LLM to read natural language problems and directly generate programs as the intermediate reasoning steps, without further processing. However, the solution step is offloaded to a runtime such as a Python interpreter. 

\section{Case Study}
\label{app:casestudy}
\input{tbls/case_studies_2}
Table \ref{tb:case2} shows an example drawn from the multiple-answer questions from TempReason L2.

\section{Prompts}
\subsection{Training Prompt for Standard Prompting of Single-Answer Questions}
\textbf{Context}: Alain Roche plays for A.J. Auxerre from Jan, 1990 to Jan, 1992. Alain Roche plays for Paris Saint-Germain F.C. from Jan, 1992 to Jan, 1998. Alain Roche plays for Valencia CF from Jan, 1998 to Jan, 2000.\\ 
\noindent\textbf{Question}: Which team did Alain Roche play for in Jan, 1995? Answer the question based on the context. Only answer the name.\\
\noindent\textbf{Answer}: Paris Saint-Germain F.C.

\subsection{Training Prompt for CoT (C+Q+R+A) Prompting of Single-Answer Questions}
\textbf{Context}: Alain Roche plays for A.J. Auxerre from Jan, 1990 to Jan, 1992. Alain Roche plays for Paris Saint-Germain F.C. from Jan, 1992 to Jan, 1998. Alain Roche plays for Valencia CF from Jan, 1998 to Jan, 2000.\\
\noindent\textbf{Question}: Which team did Alain Roche play for in Jan, 1995? Answer the question based on the context. Reason first and then answer the question. Only answer the name.\\
\noindent\textbf{Reasoning}: First, Jan, 1995 is in between Jan, 1992 and Jan, 1998. Second, Alain Roche plays for Paris Saint-Germain F.C. from Jan, 1992 to Jan, 1998.\\
\textbf{Answer}: Paris Saint-Germain F.C.

\subsection{Training Prompt for Standard Prompting of Multiple-Answer Questions}
\textbf{Context}: Alain Roche plays for A.J. Auxerre from Jan, 1990 to Jan, 1992. Alain Roche plays for Paris Saint-Germain F.C. from Jan, 1992 to Jan, 1998. Alain Roche plays for Valencia CF from Jan, 1998 to Jan, 2000. \\
\noindent\textbf{Question}: Which team did Alain Roche play for in Jan, 1992? Answer the question based on the context. Only answer the name, and separate by comma.\\
\noindent\textbf{Answer}: A.J. Auxerre, Paris Saint-Germain F.C.

\subsection{Training Prompt for CoT (Q+C+R+A) Prompting of Multiple-Answer Questions}
\noindent\textbf{Question}: Which team did Alain Roche play for in Jan, 1995? Answer the question based on the context. Reason first and then answer the question. Only answer the name, and separate by comma.\\
\textbf{Context}: Alain Roche plays for Paris Saint-Germain F.C. from Jan, 1992 to Jan, 1998. Alain Roche plays for Olympique de Marseille from Jan, 1989 to Jan, 1990. Alain Roche plays for France national association football team from Jan, 1988 to Jan, 1996.\\
\noindent\textbf{Reasoning}: First, Jan, 1995 is in between Jan, 1992 and Jan, 1998. Jan, 1995 is also in between Jan, 1988 and Jan, 1996. Second, Alain Roche plays for Paris Saint-Germain F.C. from Jan, 1992 to Jan, 1998. Alain Roche plays for France national association football team from Jan, 1988 to Jan, 1996.\\ 
\textbf{Answer}: Paris Saint-Germain F.C., France national association football team.

\subsection{Training Prompt for Structural Information Extraction}
\label{sec:sir}

Extract information from the question and context. Strictly follow the below example.
\bigbreak
\noindent \textbf{Question}: Who was the owner of Westfield Montgomery before Westfield Group?\\
\textbf{Context}: Westfield Montgomery is owned by Unibail Rodamco Westfield from Jun, 2018 to Dec, 2022. Westfield Montgomery is owned by The May Department Stores Company from Mar, 1968 to Jan, 1971. Westfield Montgomery is owned by Westfield Group from Jan, 1971 to Jan, 2014.\\
\noindent \textbf{extracted\textunderscore info} = \{(datetime(2018, 6, 1), datetime(2022, 12, 1)): ``Unibail Rodamco Westfield'', (datetime(1968, 3, 1), datetime(1971, 1, 1)): ``The May Department Stores Company'', (datetime(1971, 1, 1), datetime(2014, 1, 1)): ``Westfield Group''\}\\
\noindent \textbf{ref\textunderscore obj} = ``Westfield Group''\\
\noindent\textbf{ref\textunderscore time} = ``before''\\

%% file: tbls/l1_algo.tex
\begin{algorithm*}
    \caption{L1 Reasoning Logic}
    \label{alg:l1code}
    \begin{algorithmic}
        \Require $\textrm{Reference Time } r_t$, $\textrm{Reference Time Relation } r_r$, $\textrm{Time Interval } r_i$;
        \Ensure $\textrm{Answer } a$;
        \If{ $r_r = $ after}
            \State $a \gets r_t + r_i$
        \Else{ $r_r = $ before}
            \State $a \gets r_t - r_i$
        \EndIf    
    \end{algorithmic}
\end{algorithm*}

%% file: tbls/l2_algo.tex
\begin{algorithm*}
    \caption{L2 Reasoning Logic}
    \label{alg:l2code}
    \begin{algorithmic}
        \Require $\textrm{Reference Time } r_t$, $\textrm{Temporal Data Dictionary } D$;
        \Ensure $\textrm{List of answers } a$;
        \Procedure{time\textunderscore relation}{$st_1$,$et_1$,$st_2$,$et_2$} \Comment{Function to determine the relation of two time periods}
            \If{$et_1 \leq st_2$}
                \Return{after}
            \ElsIf{$et_2 \leq st_1$}
                \Return{before}
            \ElsIf{$st_2 < st_1$ and $et_2 > et_1$}
                \Return{contains}
            \ElsIf{$st_1 < st_2)$ and $et_1 > et_2$}
                \Return{contained by}
            \ElsIf{$st_1 = st_2$ and $et_1 = et_2$}
                \Return{simultaneous}
            \Else{}
                \Return{overlaps}
            \EndIf
        \EndProcedure

        \If{$r_t$ contains $r_{st}, r_{et}$ and $r_{st} \neq r_{et}$} \Comment{If the reference time is a time span}
            \For{$k,v$ in $D$}
                \If{\Call{time\textunderscore relation}{$r_{st}$,$r_{et}$,$k_{st}$,$k_{et}$} \textbf{is not} before \textbf{and} after}
                    \State \textbf{insert} $v$ \textbf{into} $a$
                \EndIf
            \EndFor
        \Else{} \Comment{The reference time is a timestamp}
            \For{$k,v$ in $D$}
                \If{$k_{st} \leq r_{t}$ \textbf{and} $k_{et} \geq r_t$}
                    \State \textbf{insert} $v$ \textbf{into} $a$
                \EndIf
            \EndFor
        \EndIf
    \end{algorithmic}
\end{algorithm*}

%% file: tbls/case_studies_2.tex
\begin{table*}[t!]
    \centering
    \scalebox{0.67}{
\begin{tabular}{ll}
\toprule
\textbf{Question}               & Which team did Paul Abrahams play for in Jan, 2001?                                                                                                                                                                                                                                                                                                                                                                                 \\ \hline
\textbf{Context}                & \begin{tabular}[c]{@{}l@{}}Paul Abrahams plays for Wivenhoe Town F.C. from Jan, 2004 to Jan, 2005. \\ Paul Abrahams plays for Heybridge Swifts F.C. from Jan, 2001 to Jan, 2004. \\ Paul Abrahams plays for Canvey Island F.C. from Jan, 2000 to Jan, 2001. \\ Paul Abrahams plays for Colchester United F.C. from Jan, 1996 to Jan, 1999. \\ Paul Abrahams plays for Chesham United F.C. from Jan, 2001 to Jan, 2001.\end{tabular} \\ \hline
\textbf{Gold Answer}            & \cellcolor[HTML]{FFFFFF}\textbf{Canvey Island F.C., Chesham United F.C., Heybridge Swifts F.C.}                                                                                                                                                                                                                                                                                                                                     \\ \hline
\textbf{Standard (InstructGPT)} & \cellcolor[HTML]{FFFFFF}\textcolor{myblue}{\textbf{Canvey Island F.C., Chesham United F.C., Heybridge Swifts F.C.}}                                                                                                                                                                                                                                                                                                                                     \\ \hline
\textbf{CoT (InstructGPT)}      & \begin{tabular}[c]{@{}l@{}}Answer: \textcolor{myred}{\textbf{Heybridge Swifts F.C.}} Reasoning: \\ First, Jan, 2001 is in between Jan, 2001 and Jan, 2004. \\ Second, Paul Abrahams plays for Heybridge Swifts F.C. from Jan, 2001 to Jan, 2004.\end{tabular}                                                                                                                                                                                                   \\ \hline
\textbf{Standard (GPT-4)}       & \cellcolor[HTML]{FFFFFF}\textcolor{myblue}{\textbf{Canvey Island F.C., Chesham United F.C., Heybridge Swifts F.C.}}                                                                                                                                                                                                                                                                                                                                     \\ \hline
\textbf{CoT (GPT-4)}            & \begin{tabular}[c]{@{}l@{}}Answer: \textcolor{myred}{\textbf{Heybridge Swifts F.C.}} Reason: \\ Jan, 2001 is in between Jan, 2001 to Jan, 2004, when Paul Abrahams played for Heybridge Swifts F.C.\end{tabular}                                                                                                                                                                                                                                                \\ \hline
\textbf{TempLogic}                   & \cellcolor[HTML]{FFFFFF}\textcolor{myblue}{\textbf{Canvey Island F.C., Chesham United F.C., Heybridge Swifts F.C.}}                                                                                                                             \\ \bottomrule
\end{tabular}
    }
    \caption{A case study of a L2 multiple-answer question from TempReason. Answers highlighted in blue are correct, whereas thoses marked in red are incorrect.}
    \label{tb:case2}
\end{table*}